\documentclass[letterpaper]{article} 
\usepackage{aaai2026}  
\usepackage{times}  
\usepackage{helvet}  
\usepackage{courier}  
\usepackage[hyphens]{url}  
\usepackage{graphicx} 
\usepackage{natbib}  
\usepackage{caption} 
\usepackage{amssymb}
\usepackage{xcolor}
\usepackage{amsmath}
\usepackage{booktabs}
\usepackage{multirow}
\usepackage{algorithm}
\usepackage{algorithmic}

\usepackage{newfloat}
\usepackage{listings}
\DeclareCaptionStyle{ruled}{labelfont=normalfont,labelsep=colon,strut=off} 
\floatstyle{ruled}
\newfloat{listing}{tb}{lst}{}
\floatname{listing}{Listing}
\title{Semantic-aware DropSplat: Adaptive Pruning of Redundant Gaussians for 3D Aerial-View Segmentation}
\title{Semantic-aware DropSplat: Adaptive Pruning of Redundant Gaussians for 3D Aerial-View Segmentation}

\title{Semantic-aware DropSplat: Adaptive Pruning of Redundant Gaussians for 3D Aerial-View Segmentation}

\author{
    Xu Tang\textsuperscript{\rm 1*},
    Junan Jia\textsuperscript{\rm 1},
    Yijing Wang\textsuperscript{\rm 1},
    Jingjing Ma\textsuperscript{\rm 1},
    Xiangrong Zhang\textsuperscript{\rm 1}
}
\affiliations{
    \textsuperscript{\rm 1}School of Artificial Intelligence,\\
    Xidian University, Xi’an, China\\
    \texttt{
        tangxu128@gmail.com, 
        24171214056@stu.xidian.edu.cn, 
        yijingwang@stu.xidian.edu.cn,\\
        smallpig32@sina.com, 
        xrzhang@mail.xidian.edu.cn
    }
}

\usepackage{bibentry}

\begin{document}

\maketitle

\begin{abstract}
In the task of 3D Aerial-view Scene Semantic Segmentation (3D-AVS-SS), traditional methods struggle to address semantic ambiguity caused by scale variations and structural occlusions in aerial images. This limits their segmentation accuracy and consistency. To tackle these challenges, we propose a novel 3D-AVS-SS approach named SAD-Splat. Our method introduces a Gaussian point drop module, which integrates semantic confidence estimation with a learnable sparsity mechanism based on the Hard Concrete distribution. This module effectively eliminates redundant and semantically ambiguous Gaussian points, enhancing both segmentation performance and representation compactness. Furthermore, SAD-Splat incorporates a high-confidence pseudo-label generation pipeline. It leverages 2D foundation models to enhance supervision when ground-truth labels are limited, thereby further improving segmentation accuracy. To advance research in this domain, we introduce a challenging benchmark dataset: 3D Aerial Semantic (3D-AS), which encompasses diverse real-world aerial scenes with sparse annotations. Experimental results demonstrate that SAD-Splat achieves an excellent balance between segmentation accuracy and representation compactness. It offers an efficient and scalable solution for 3D aerial scene understanding.
\end{abstract}

\section{Introduction}


As a branch of 3D multi-view semantic segmentation, 3D Aerial-view Scene Semantic Segmentation (3D-AVS-SS) plays a crucial role in various remote sensing (RS) applications, including land use monitoring, urban planning, and disaster response \cite{huang2023deep, rahnemoonfar2022rescuenet}. Its goal is to assign semantic labels to each pixel in multi-view aerial images captured from high-altitude perspectives around a target scene. Compared to the normal 2D images, aerial images often suffer from significant scale variations and structural occlusions, which hinder spatial consistency and semantic alignment across multiple views. Therefore, constructing a unified 3D scene representation and performing semantic reasoning in 3D multi-view aerial images has become a promising research topic for achieving more accurate and consistent segmentation results.

There are two steps in the standard 3D multi-view semantic segmentation process. The first step is to reconstruct the 3D scene using the multi-view images. The second step is to embed the semantic features into the obtained 3D scene and render the semantic 3D scene back into 2D views for semantic predictions. In the 3D reconstruction step, Neural Radiance Fields (NeRF) \cite{mildenhall2021nerf} and 3D Gaussian Splatting (3DGS) \cite{kerbl20233d} are two popular tools that have gained significant attention in recent years. NeRF introduces an implicit volumetric representation by learning density and color fields from multi-view images, enabling photorealistic rendering. In contrast, 3DGS employs an explicit and structured modeling paradigm based on parameterized 3D Gaussians, which jointly encode geometry and appearance. It is gaining increasing attention due to its rendering efficiency and structural interpretability. For this reason, our work is also built upon 3DGS. For the semantic segmentation step, powerful 2D foundation models (e.g., CLIP \cite{radford2021learning} and SAM \cite{kirillov2023segment}) are first leveraged to extract semantic knowledge, which is subsequently distilled into the 3D representation. Semantic segmentation is then performed by rendering the 3D scene onto 2D images. Based on the basic techniques mentioned above, many successful 3D multi-view semantic segmentation models have been developed \cite{zhou2024feature, cen2025segment}.

\begin{figure*}[!htpb]
  \centering
  \includegraphics[width= \textwidth]{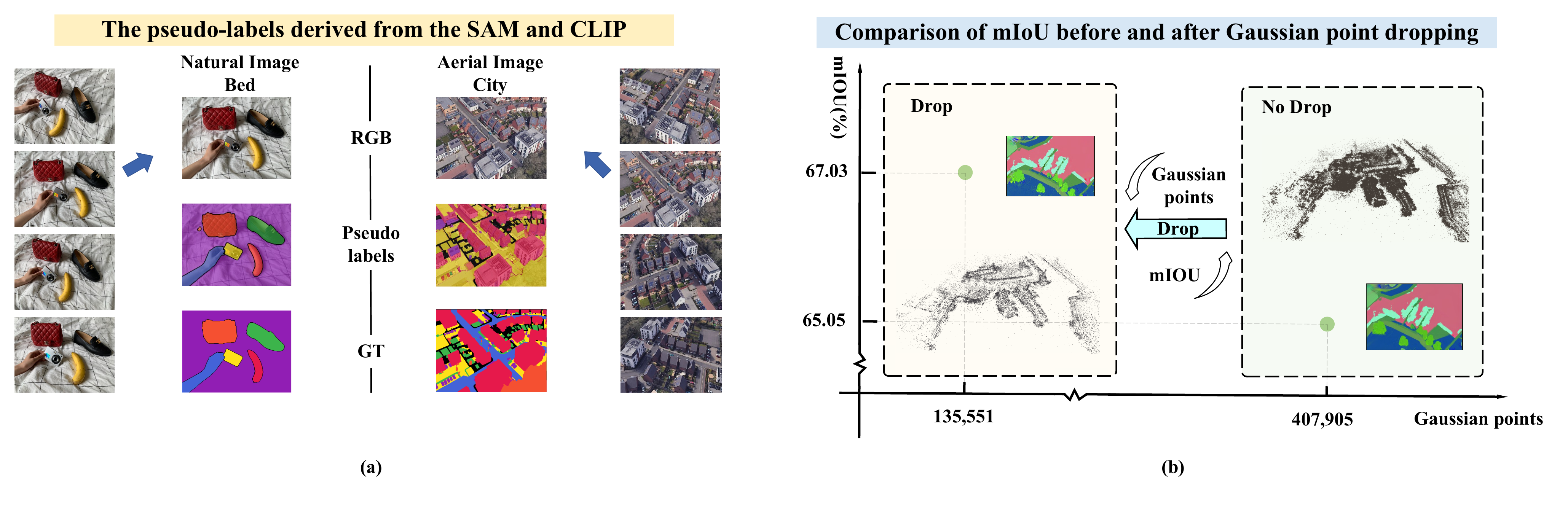}
  \caption{ (a) Comparison of segmentation results between natural and aerial images using foundation models.  (b) Effectiveness of the semantic-aware drop. The proposed drop mechanism significantly reduces the number of Gaussian points while maintaining segmentation accuracy, leading to a more compact and efficient model representation.}
  \label{fig1:introduction}
\end{figure*}

Although the above methods perform well, applying them to 3D-AVS-SS tasks directly still presents many challenges. In the 3D reconstruction step, since manual annotation of aerial images is scarce, semantically ambiguous regions often appear. Within the 3DGS framework, the model tends to frequently duplicate or split Gaussian points in uncertain areas, producing dense, redundant, and noisy semantic representations. This ultimately undermines the spatial consistency and reduces the overall compactness of the results. In the semantic segmentation step, as the prevalent 2D foundation models are always pre-trained by natural images, the semantic knowledge transferred from them for multi-view aerial images often contains noise and inconsistencies, which lead to the problems of category confusion, misclassification, and inter-view disagreement. Taking Fig.~\ref{fig1:introduction}a as an example, the left side presents a natural scene (``bed’’) \cite{liu2023weakly} with its corresponding RGB image, ground truth (GT) segmentation, and the prediction from a 2D foundation model. It can be observed that the model performs reasonably well in this setting. In contrast, the right side shows an example from an aerial scene (``city’’). Compared to the GT, the prediction from the 2D foundation model contains numerous errors; for example, roads and trees are rarely correctly classified, resulting in a large amount of noisy supervision.

To address the above challenges, we propose a new 3D-AVS-SS model named Semantic-aware DropSplat (SAD-Splat) under the framework of 3DGS. To tackle the issue of excessive redundant and semantically ambiguous Gaussian points generated during the 3D reconstruction process in 3DGS, SAD-Splat introduces a learnable drop strategy. This strategy estimates the retention probability of each Gaussian point by integrating a base drop rate, semantic confidence estimation, and a sparsity constraint based on the hard-concrete \cite{louizos2017learning} distribution. Also, it effectively suppresses Gaussian points that are semantically ambiguous or contribute little, resulting in a more efficient and compact model representation, the effect of applying this strategy to the same scene is shown in Fig. 1b. In addition to tackling the dissatisfactory performance of 2D foundation models on multi-view aerial images, SAD-Splat incorporates a filtering mechanism that only utilizes semantic information with sufficiently high confidence for supervision, thereby minimizing the influence of semantically uncertain regions.

To advance research in 3D-AVS-SS, we construct a new dataset named 3D-Aerial Semantic (3D-AS). It contains diverse real-world aerial scenes with pixel-level annotations and poses challenges such as limited supervision, class imbalance, and scene diversity.
\section{Related Work}

In this work, our core pipeline leverages pre-trained foundation models to extract semantic features from images, which are subsequently transferred to 3D space via knowledge distillation. Based on this technical framework, we review related work from two primary perspectives. First, we discuss foundation models widely employed in image segmentation tasks, encompassing vision-language models and large-scale segmentation networks. Second, we summarize recent advancements in embedding semantic features within the 3D Gaussian Splatting (3DGS) framework.

\noindent\textbf{Foundation Models for Segmentation}
In recent years, the development of foundation models has significantly advanced semantic segmentation. These models can be broadly categorized into two groups. The first group comprises vision-language models. CLIP achieves open-vocabulary recognition by learning joint image-text embeddings from large-scale datasets; OpenCLIP~ \cite{cherti2023reproducible}, as an optimized open-source version, enhances training efficiency and applicability. GeoRSCLIP ~\cite{zhang2024rs5m} aligns aerial images with geographic semantics through contrastive learning, demonstrating superior performance in land cover classification and scene understanding. The second group focuses on segmentation models, such as DINO \cite{caron2021emerging} and CLIP-based approaches like MaskCLIP ~\cite{dong2023maskclip} and LSeg ~\cite{li2022language}, which integrate cross-modal semantic understanding with dense spatial prediction. Additionally, segmentation models based on the SAM and its latest version, SAM2 ~\cite{ravi2024sam}, trained on massive datasets, exhibit strong generalization and zero-shot capabilities.

However, these models still face limitations in 3D-AVS-SS. Insufficient exploitation of geometric relationships across views and domain gaps results in degraded performance.

\noindent\textbf{Semantic Segmentation with 3D Gaussian Splatting}
Recent work, such as feature-3DGS~\cite{zhou2024feature}, distills features from large 2D foundation models into 3DGS, enabling semantic rendering and open-vocabulary segmentation. LangSplat~\cite{qin2024langsplat} introduces language supervision to build 3D semantic fields that support precise spatial queries. OpenSplat3D~\cite{piekenbrinck2025opensplat3d} associates semantic information with Gaussian points, leveraging SAM masks and contrastive loss combined with vision-language embeddings to achieve open-vocabulary 3D instance segmentation. This approach demonstrates strong cross-view consistency and high-accuracy segmentation across multiple datasets, significantly advancing the application of 3DGS for semantic segmentation.

Despite these advances, existing methods still face challenges such as oversaturation of Gaussian points and semantic confusion. To address these issues, we propose a semantic-aware drop mechanism within the 3DGS framework. This mechanism jointly prunes redundant Gaussian points and enhances semantic compactness, thus improving the 3D-AVS-SS performance.
\section{Method}

\begin{figure*}[!ht]
  \centering
  \includegraphics[width=\textwidth, height=0.4\textheight, keepaspectratio]{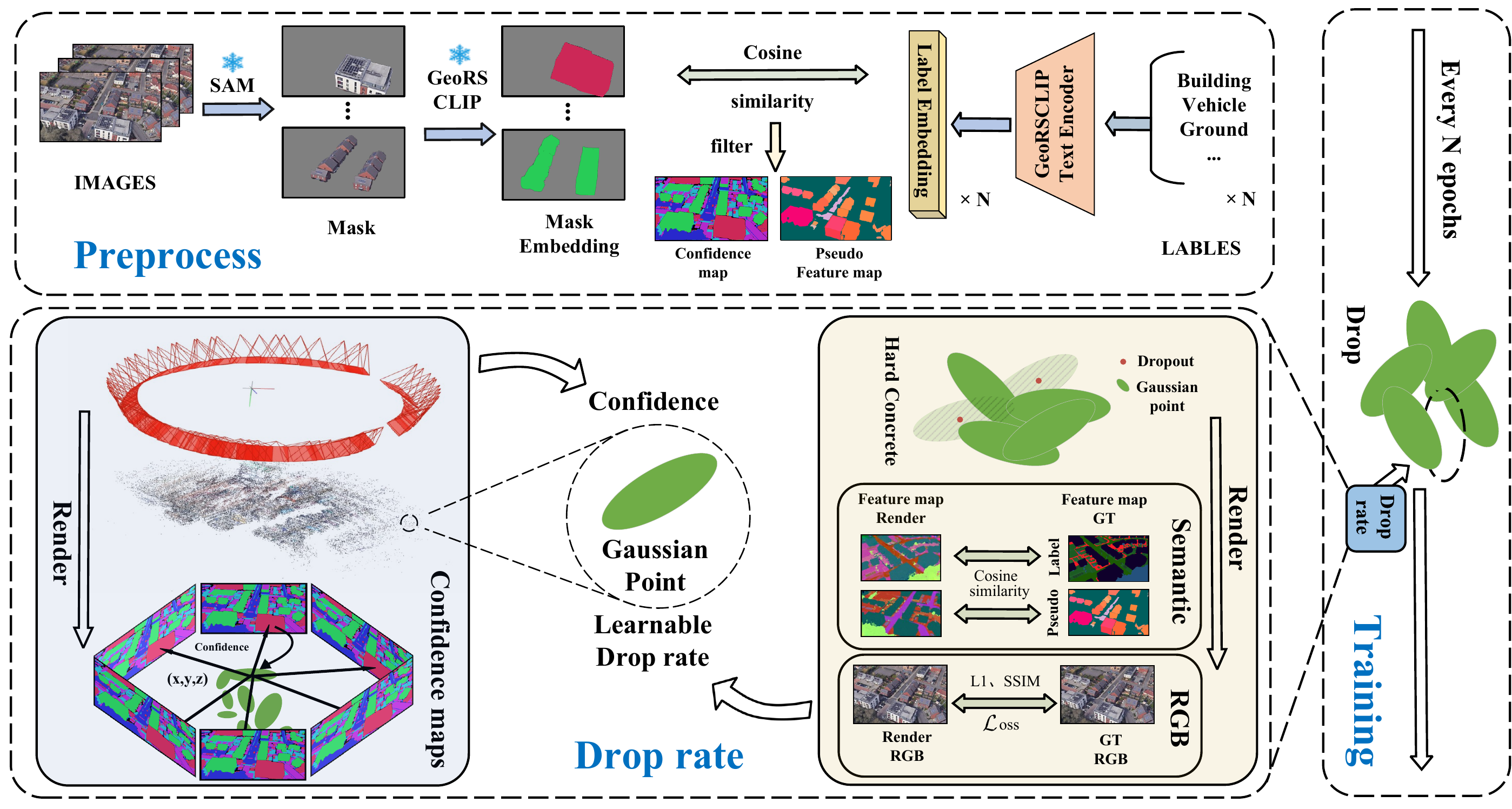}
  \caption{Overview of SAD-Splat. The upper part shows the preprocessing pipeline, where SAM, GeoRSCLIP, and a high-confidence filtering strategy are used to generate semantic feature maps and confidence maps. The right part depicts the training process, in which a drop operation is performed after a fixed number of training iterations and is repeated throughout training. The drop operation is guided by the drop rate produced by the bottom module, which integrates a base drop rate, semantic confidence, and a learnable drop parameter to estimate a drop probability for each Gaussian point.}
  \label{fig2:method}
\end{figure*}

The overall pipeline of SAD-Splat is illustrated in Fig.~\ref{fig2:method}. Given multi-view aerial images, a small number of ground-truth semantic labels, and corresponding text descriptions, the preprocessing stage generates high-confidence pseudo-labels and semantic confidence maps. During training, the system jointly reconstructs semantic features and learns both the semantic confidence and the learnable drop rate for each Gaussian point. This joint optimization integrates semantic importance and structural redundancy, ensuring both semantic accuracy and model compactness. 

\subsection{Preliminary}

\textbf{3DGS} \cite{kerbl20233d} is an efficient method for representing complex 3D scenes using a collection of anisotropic Gaussian primitives. Each Gaussian simultaneously encodes both geometric and appearance-related information, including spatial position, color, opacity, rotation, and scale, enabling a unified representation of scene structure and visual content.

Given a set of multi-view images with camera poses, 3DGS initializes a set of Gaussians as follows:
\[
G = \{ {g_i}\} _{i = 1}^N,
\]
where each Gaussian $g_i$ is defined as a five-tuple:
\[
{g_i} = \{ {\mu _i},{c_i},{\alpha _i},{R_i},{s_i}\}, 
\]
where $\mu_i \in \mathbb{R}^3$ denotes the 3D center, $c_i \in \mathbb{R}^3$ represents the RGB color, $\alpha_i \in [0,1]$ indicates the opacity, $R_i \in \mathbb{R}^{3 \times 3}$ implies the rotation matrix, and $s_i \in \mathbb{R}^3$ refers to the anisotropic scaling factors.During the rendering process, Gaussians are projected onto the image plane and undergo front-to-back alpha compositing based on view-dependent opacity to achieve proper visibility ordering.

Based on 3DGS, Feature 3DGS~\cite{zhou2024feature} introduces a learnable semantic embedding vector $f_i \in \mathbb{R}^d$ for each Gaussian, enabling dense semantic modeling in 3D space. These semantic features are aligned with a 2D foundation model through distillation to obtain cross-modal supervision signals. In the semantic rendering stage, the semantic descriptor $S(p)$ for a pixel $p$ is computed by weighted aggregation of the semantic features of its visible Gaussian set $g_p$, where both opacity and visibility factors determine the weights. Through this mechanism, semantic information is effectively incorporated into the 3DGS framework, enabling semantic segmentation of 3D scenes based on Gaussian representations. Our SAD-Splat is developed based on Feature 3DGS. 

\subsection{Gaussian Point Drop Implementation}

During the \textbf{\textit{training stage}} of Feature 3DGS, regions with semantic ambiguity or conflict often accumulate large gradients. These large gradients can trigger density control operations, such as splitting or cloning, which may lead to the generation of redundant and semantically ambiguous regions. To address this issue, we propose a drop operation specifically targeting these ambiguous regions to enhance the efficiency and stability of SAD-Splat. The drop operation is designed by comprehensively considering both semantic and structural factors.

\noindent\textbf{Semantic Confidence Drop Module}

To evaluate the semantic importance of Gaussian points in 3DGS, we generate a confidence map using SAM ~\cite{kirillov2023segment} and GeoRSCLIP ~\cite{zhang2024rs5m}. SAM segments training images to extract object regions and their edge information. Meanwhile, GeoRSCLIP’s image encoder derives semantic feature vectors for objects, and its text encoder extracts semantic vectors for category descriptions. Cosine similarities between these vectors yield per-category similarity scores, which are mapped to image coordinates to form a confidence map \(\mathbf{Confidence} \in \mathbb{R}^{H \times W \times C}\), where \(H\) and \(W\) denote image height and width, and \(C\) means the number of categories. For each rendered image, the 3D centers of visible Gaussian points are projected onto the confidence map, and the maximum score across the category dimension, weighted by the point’s opacity, is extracted as the view-specific confidence. For multi-view visible points, the average confidence across views is computed to assess their semantic importance in the scene. Semantically ambiguous regions typically exhibit lower confidence scores.

\noindent\textbf{Learnable Structure Drop Module}

To identify Gaussian points in the model structure that can be dropped, we utilize the Hard Concrete distribution ~\citep{louizos2017learning} to obtain a differentiable approximation of binary drop, thereby optimizing Gaussian point pruning in 3DGS. Each Gaussian point is assigned a learnable parameter \(\log \alpha\), which governs the drop probability in logarithmic form to facilitate gradient-based optimization. A binary mask is sampled using noise-augmented logits and a bounded sigmoid function, enabling end-to-end training. To promote sparsity, we introduce an \(\mathcal{L}_0\) regularization term to penalize the expected number of active Gaussian points. The non-zero activation probability is defined as:
\[
P_{\mathrm{nonzero}} = \sigma\left( \frac{\log \alpha - \tau \cdot \log\left( \frac{\mathrm{threshold}}{1 - \mathrm{threshold}} \right)}{\tau} \right),
\]
where \(\log \alpha\) is the learnable parameter controlling retention, \(\tau\) is the temperature parameter adjusting distribution smoothness, and \(\mathrm{threshold} = \frac{-l}{r - l}\) is derived from Hard Concrete bounds \(l\) and \(r\). Then, the regularization loss can be defined as:
\[
\mathcal{L}_{L_0} = \mathbb{E}[P_{\mathrm{nonzero}}].
\]
It minimizes the expected non-zero probability, effectively reducing redundant Gaussian points.

After $N$ training iterations, we integrate the base drop rate \(P_{\mathrm{base}}\), the average semantic confidence \(P_{\mathrm{confidence}}\), and the learnable drop rate \(P_{\mathrm{learned}}\) to compute the comprehensive drop probability for each Gaussian point
\[
P_{\mathrm{drop}} = P_{\mathrm{base}} \cdot (1 - P_{\mathrm{confidence}}) \cdot P_{\mathrm{learned}}.
\]
Here, \(P_{\mathrm{base}}\) controls the overall drop scale, \(P_{\mathrm{confidence}}\) reflects semantic importance, and \(P_{\mathrm{learned}}\) enables adaptive adjustment. Periodic pruning is performed every $N$ iterations to remove semantically ambiguous or structurally invalid Gaussian points, effectively suppressing redundant regions.

\subsection{Pseudo-Label Generation Module}

In 3D-AVS-SS tasks, the absence of ground-truth labels leads to inadequate supervision. This challenge is particularly evident for views where annotations for targets, such as buildings, are missing. To address this issue, we introduce a \textbf{\textit{preprocess stage}} where 2D foundation models are leveraged to generate pseudo-labels, providing auxiliary supervision signals.

We reuse the confidence map generation process to compute similarity scores between the image-derived features of object regions and the text-derived features of category descriptions. To ensure the reliability of supervision signals, three metrics are computed for each target: the TOP1 score (the similarity with the most likely class), the difference between TOP1 and TOP2 scores (denoted as \(\Delta TOP\), represents the model’s confidence dominance in favor of the predicted category), and entropy (indicating the uncertainty of the semantic distribution). The mean and standard deviation for each metric are calculated over all target samples in the training dataset to reflect the current data distribution. Only targets satisfying the following conditions are retained:
\[
\begin{cases}
TOP1 > mean(TOP1) + std(TOP1), \\
\Delta TOP > mean(\Delta TOP) + std(\Delta TOP), \\
Entropy < mean(Entropy) + std(Entropy).
\end{cases}
\]
These conditions filter out targets with low confidence or high ambiguity, ensuring robust pseudo-labels. The features of selected targets are embedded back into their original image positions to construct a pseudo-feature map. Non-target regions are assigned zero values and masked to exclude them from gradient computations, thereby focusing optimization on reliable semantic regions.

\subsection{Training Loss}

The training loss comprises a semantic feature loss and an RGB reconstruction loss. The semantic feature loss encourages alignment between the predicted feature map and the ground-truth feature map by maximizing the cosine similarity at each pixel. It is defined as:
\[
\mathcal{L}_\mathrm{semantic} = 
1 - \frac{1}{|\Omega|} \sum_{(x,y)\in\Omega}
\frac{f_\mathrm{pred}(x,y) \cdot f_\mathrm{gt}(x,y)}
{\|f_\mathrm{pred}(x,y)\|_2 \cdot \|f_\mathrm{gt}(x,y)\|_2},
\]
where $f_{pred}$ is the predicted feature at pixel $(x,y)$, $f_{gt}$ is the ground-truth feature, and $\Omega$ denotes the set of all pixel positions.The RGB reconstruction loss, following 3D Gaussian Splatting practices, combines L1 and SSIM terms to balance pixel-level accuracy and perceptual quality:
\[
\begin{aligned}
\mathcal{L}_\mathrm{rgb} =\ & (1 - \lambda_\mathrm{SSIM}) \cdot \|I_\mathrm{rendered} - I_\mathrm{input}\|_1 \\
& + \lambda_\mathrm{SSIM} \cdot \left(1 - \mathrm{SSIM}(I_\mathrm{rendered}, I_\mathrm{input})\right).
\end{aligned}
\]
Then, the total loss function is formulated as:
\[
\mathcal{L} = \lambda_\mathrm{semantic} \cdot \mathcal{L}_\mathrm{semantic} + \lambda_\mathrm{rgb} \cdot \mathcal{L}_\mathrm{rgb} + \lambda_{L_0} \cdot \mathcal{L}_{L_0},
\]
where $\lambda_\mathrm{semantic}$, $\lambda_\mathrm{rgb}$, and $\lambda_{L_0}$ are weighting coefficients, and $\mathcal{L}_{L_0}$ represents the regularization loss introduced by the Gaussian point drop module.
 \section{Dataset}
To evaluate  SAD-Splat, we constructed a highly challenging dataset,  named 3D-AS. The dataset comprises three scene categories sourced from Google Earth, with each category containing three scenes, resulting in a total of nine real-world sub-scenes.

To simulate surround-view capture around target scenes, we extracted screenshots from Google Earth. For each scene, approximately 70 images were captured at roughly equal angular intervals, covering diverse viewpoints, with each image having a resolution of roughly 1600 × 900 pixels. By leveraging Google Earth’s scale reference and the size characteristics of common objects, we estimated that each pixel corresponds to approximately 0.3–0.5 meters in the physical world. Each scene shares the same set of pixel-level label categories, covering six or more common object types. For every sub-scene, 10 images were manually annotated. Of these, 10\% (7 images) were designated as the test set, with the remaining 3 images used for training. An overview of the dataset is provided in Table~\ref{tab:3d_as_details}, while detailed information for each scene will be included in the supplementary material.

\begin{figure}[!ht]
  \centering
  \includegraphics[width=\columnwidth]{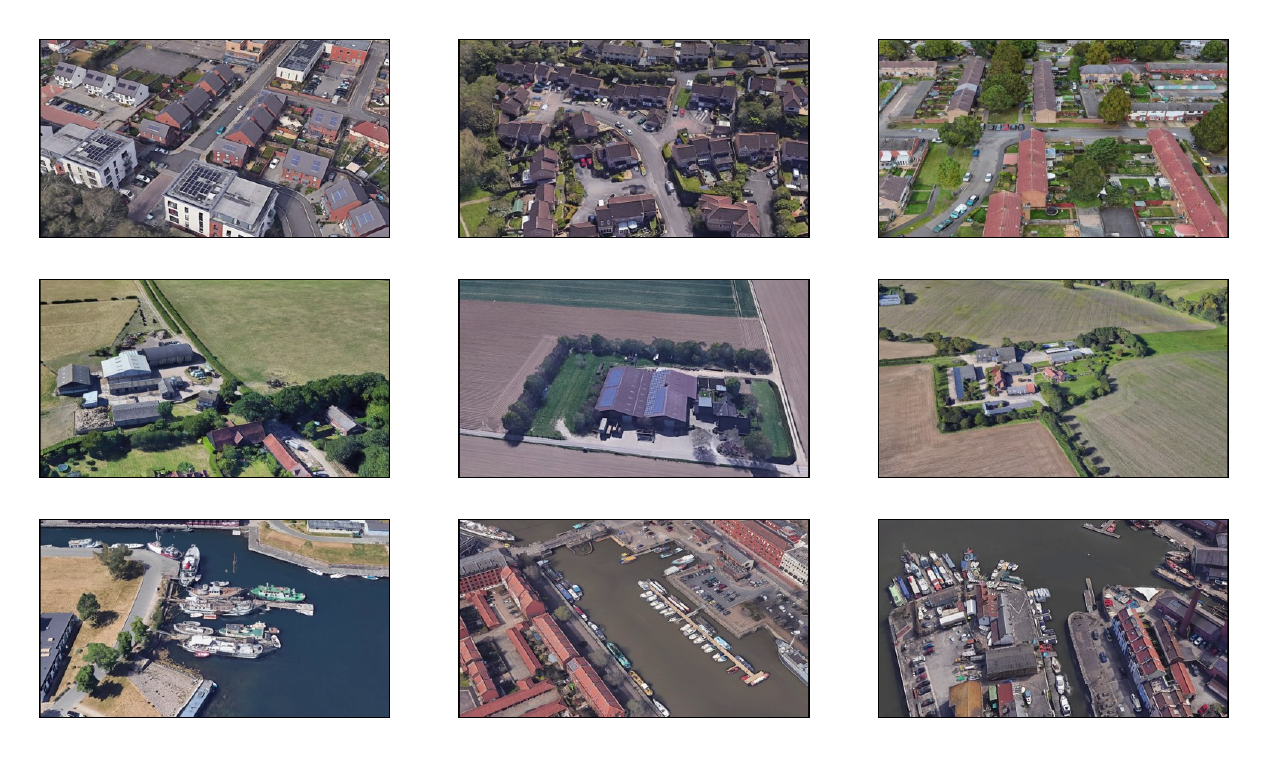}
  \caption{Displays the first image of each sub-dataset in the dataset.}
  \label{fig3:experiment1}
\end{figure}

3D-AS presents three key challenges. First, labeled samples are scarce, with only 4.8\% (3/63) of training samples annotated. Second, significant intra-class variance and inter-class imbalance exist; for instance, the ``Building'' label encompasses diverse architectural types, while images often contain abundant ``Ground'' and ``Building'' labels but minimal ``Grass'' in Fig.~\ref{fig3:experiment1}. Third, the dataset includes varied real-world environments, specifically City (dense urban areas with complex structures), Country (open rural fields with sparse objects), and Port (waterfront regions with reflective surfaces and industrial facilities), each presenting distinct spatial and semantic distributions that pose significant challenges for model generalization.

\begin{table}[!t]
  \centering
  \begin{tabular}{l c c c}
    \toprule
    Scene & Views & Categories & Effective Pixel(\%) \\
    \midrule
    City \#0     & 70 & 6 & 89.0 \\
    City \#1     & 70 & 6 & 91.0 \\
    City \#2     & 70 & 6 & 91.8 \\
    Country \#0  & 70 & 7 & 95.9 \\
    Country \#1  & 70 & 7 & 98.6 \\
    Country \#2  & 70 & 7 & 99.8 \\
    Port \#0     & 70 & 8 & 99.8 \\
    Port \#1     & 70 & 8 & 97.9 \\
    Port \#2     & 70 & 8 & 98.8 \\
    \bottomrule
  \end{tabular}
  \caption{Overview of 3D-AS Dataset}
  \label{tab:3d_as_details}
\end{table}

\section{Experiments} 

In this section, we present the basic experimental results and analysis of SAD-Splat. 

\begin{table*}[ht!]
\centering
\resizebox{\textwidth}{!}{%
\begin{tabular}{l|cc|cc|cc|cc|cc|cc|cc|cc|cc}
\toprule
\multicolumn{1}{c|}{\multirow{2}{*}{\textbf{Method}}} & \multicolumn{6}{c|}{\textbf{City}} & \multicolumn{6}{c|}{\textbf{Country}} & \multicolumn{6}{c}{\textbf{Port}} \\
\cmidrule(r){2-7} \cmidrule(r){8-13} \cmidrule(r){14-19}
& \multicolumn{2}{c|}{Scene 0} & \multicolumn{2}{c|}{Scene 1} & \multicolumn{2}{c|}{Scene 2} & \multicolumn{2}{c|}{Scene 0} & \multicolumn{2}{c|}{Scene 1} & \multicolumn{2}{c|}{Scene 2} & \multicolumn{2}{c|}{Scene 0} & \multicolumn{2}{c|}{Scene 1} & \multicolumn{2}{c}{Scene 2} \\
& mIoU & mAcc & mIoU & mAcc & mIoU & mAcc & mIoU & mAcc & mIoU & mAcc & mIoU & mAcc & mIoU & mAcc & mIoU & mAcc & mIoU & mAcc \\
\midrule
LSeg & 37.5 & 67.0 & 41.1 & 74.9 & 44.5 & 73.2 & 22.5 & 31.4 & 25.4 & 27.2 & 17.6 & 21.5 & 45.2 & 77.0 & 28.2 & 55.8 & 24.0 & 62.0  \\
MaskCLIP & 33.1 & 49.5 & 28.8 & 44.9 & 37.8 & 55.0 & 23.8 & 33.6 & 30.2 & 52.0 & 21.5 & 41.6 & 46.9 & 76.0 & 35.6 & 63.4 & 31.8 & 67.2  \\
LERF & 11.7 & 34.5 & 7.2 & 23.3 & 7.5 & 23.8 & 5.5 & 20.2 & 4.0 & 16.3 & 4.1 & 17.6 & 4.5 & 14.7 & 7.6 & 26.5 & 4.6 & 17.7  \\
LangSplat & 11.2 & 26.6 & 8.3 & 20.3 & 1.7 & 4.1 & 5.4 & 12.7 & 3.0 & 11.6 & 2.3 & 9.2 & 5.3 & 10.3 & 4.1 & 9.2 & 2.7 & 5.9  \\
Feature 3DGS & 10.6 & 28.6 & 31.5 & 54.7 & 35.0 & 64.6 & 27.3 & 45.9 & 40.1 & 64.5 & 31.0 & 71.2 & 41.4 & 70.2 & 27.4 & 49.4 & 26.9 & 58.3 \\
Gaussian Grouping  & 42.9 & 69.7 & 19.4 & 46.2 & 23.1 & 47.4 & 40.6 & 81.5 & 50.0 & 88.5 & 39.2 & 86.2 & 55.3 & 85.5 & 36.6 & 65.3 & 27.9 & 68.2  \\
\midrule
\textbf{SAD-Splat (Ours)} & \textbf{69.2} & \textbf{85.7} & \textbf{65.2} & \textbf{84.0} & \textbf{61.2} & \textbf{78.6} & \textbf{56.8} & \textbf{87.7} & \textbf{69.2} & \textbf{95.1} & \textbf{53.8} & \textbf{91.4} & \textbf{67.2} & \textbf{91.7} & \textbf{57.1} & \textbf{84.7} & \textbf{47.4} & \textbf{90.8} \\
\bottomrule
\end{tabular}
}
\caption{Quantitative comparisons with state-of-the-art methods across various scene types are presented. Metrics, including mean Intersection over Union (mIoU) and mean Accuracy (mAcc), are reported in percentages (\%). The highest scores are highlighted in bold.}
\label{tab:main_results}
\end{table*}
\begin{figure*}[!ht]
  \centering
  \includegraphics[width=\textwidth, height=\textheight, keepaspectratio]{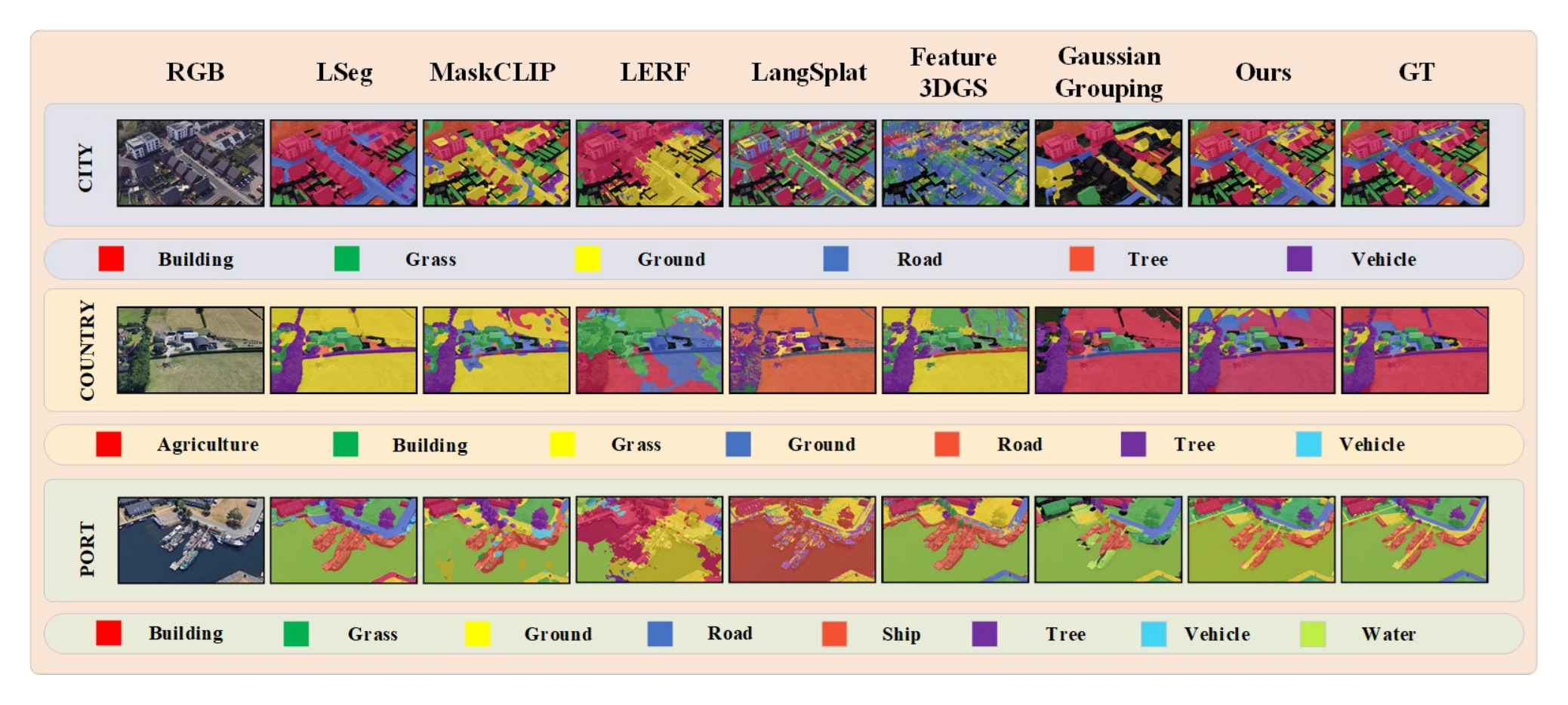}
  \caption{Visual comparison of segmentation results produced by different methods on the City and Country datasets.}
  \label{fig3:vision}
\end{figure*}

\subsection{Experimental Setting}

All experiments were conducted on 3D-AS. To ensure reproducibility, we fixed the random seed to 42 and accordingly split the dataset into training and test sets. Specifically, images numbered 0, 7, and 28 were used as the training set, while images numbered 14, 21, 35, 42, 49, 56, and 63 were used as the test set. Performance evaluation metrics include mean Intersection over Union (mIoU) and accuracy.

\noindent \textbf{Implementation Details:} 

All experiments were conducted on an NVIDIA GeForce RTX 3090 GPU with 24GB of memory and PyTorch. During training, the dimensions of the images were uniformly reduced to half the width and height of the original images. We conducted experiments with 7000 training epochs across all scenes, setting the hyperparameter for the number of discarded epochs $ N $ to 500. Our method simultaneously reconstructs RGB and semantic information, with the semantic dimension set to 32.

\subsection{Comparison Experiments}

To evaluate the effectiveness of SAD-Splat, we conducted a comprehensive comparison with state-of-the-art methods in two categories: (1) 2D open-vocabulary segmentation methods, including LSeg ~\cite{kerr2023lerf} and MaskCLIP ~\cite{fu2024featup}; (2) 3D open-vocabulary segmentation methods, including LERF ~\cite{kerr2023lerf}, LangSplat ~\cite{qin2024langsplat}, and Feature 3DGS ~\cite{zhou2024feature}. All methods were evaluated on our proposed 3D-AS.

Quantitative results are presented in Table~\ref{tab:main_results}, with visualizations shown in Fig.~\ref{fig3:vision}. Although 2D methods exhibit competitive performance in certain scenes, their lack of explicit 3D structure modeling often leads to poor spatial consistency across multiple views. 3D open-vocabulary methods show unstable performance in complex scenes with semantic ambiguity. In contrast, SAD-Splat achieves the best overall performance across all types of scenes, demonstrating its superiority in balancing high segmentation accuracy and compact representation. Its capabilities in multi-view semantic integration and structure-aware sparse modeling make it an efficient and scalable solution for the 3D-AVS-SS task.

\subsection{Ablation Study}

To evaluate the individual contributions of modules in SAD-Splat, a series of ablation experiments was designed. SAD-Splat comprises two primary components: the Gaussian Point Drop module and the Pseudo-Label Generation module. The Gaussian Point Drop module integrates learnable structural validity and semantic confidence-guided mechanisms to filter 3D Gaussian points during training dynamically, enabling adaptive compression of the model structure. The Pseudo-Label Generation module leverages the feature extraction and semantic matching capabilities of SAM and GeoRSCLIP to generate high-quality pseudo-labels for unlabeled images, thereby enhancing the coverage and diversity of supervision signals.

To validate the effectiveness of these components, the following comparative experimental setups were constructed under the City scene:

\begin{itemize}
    \item \textbf{\textit{Net-1}}: Base model with 3 GT supervision (Baseline).
    \item \textbf{\textit{Net-2}}: Baseline with Gaussian Point Drop module ($P_{\text{base}} = 1.0$).
    \item \textbf{\textit{Net-3}}: Baseline with Pseudo-Label Generation module (generating pseudo-labels for the remaining 60 unlabeled images).
    \item \textbf{\textit{Net-4}}: Baseline with both Gaussian Point Drop and Pseudo-Label Generation modules (complete SAD-Splat).
\end{itemize}

\begin{table}[t!]
\centering
\begin{tabular}{lcc}
\toprule
\textbf{Network} & \textbf{mIoU (\%)} & \textbf{Accuracy (\%)} \\
\midrule
Net-1 & 64.44 & 82.01 \\
Net-2 & 64.95 & 82.68 \\
Net-3 & 64.68 & 82.10 \\
Net-4 & \textbf{65.22} &  \textbf{82.74} \\
\bottomrule
\end{tabular}
\caption{Performance comparison across various network configurations is presented. The top-performing results are highlighted in bold.}
\label{tab:ablation_results}
\end{table}

Table~\ref{tab:ablation_results} presents the performance comparison of each configuration. Here, ``the base model'' refers to a 3D semantic segmentation framework built upon Feature 3DGS and trained with ground-truth segmentation labels from three viewpoints (3 GT supervision). Net-1 serves as the baseline, achieving an mIoU of 64.44\% and an accuracy of 82.01\%. Net-2 incorporates the Gaussian Point Drop module, guided by semantic confidence and learnable structural validity, resulting in improved performance with an mIoU of 64.95\% and an accuracy of 82.68\%. This demonstrates that selectively removing redundant or inefficient Gaussian points enhances representational capacity. Net-3 integrates the Pseudo-Label Generation module, which provides semantic pseudo-supervision for unlabeled images, yielding an mIoU of 64.68\% and an accuracy of 82.10\%. This highlights the effectiveness of additional supervision signals in improving model generalization. Net-4 combines both modules, achieving the best performance with an mIoU of 65.22\% and an accuracy of 82.74\%, confirming the synergistic effect of the modules in SAD-Splat for enhancing 3D semantic understanding.

\textbf{Analysis of the Contribution of Gaussian Point Drop submodules}

\begin{table}[t!]
\centering
\begin{tabular}{lcc}
\toprule
\textbf{Network} & \textbf{mIoU (\%)} & \textbf{Accuracy (\%)} \\
\midrule
SAD-Splat & \textbf{65.22} & \textbf{82.74} \\
Semantic Confidence Drop & 63.79 & 82.57 \\
Learnable Structure Drop & 64.01 & 82.70 \\
\bottomrule
\end{tabular}
\caption{Impact of Various Drop Strategies on Experimental Performance. Best results are highlighted in bold.}
\label{tab:dropresults}
\end{table}

To assess the contributions of the two submodules in SAD-Splat's Gaussian Point Drop mechanism, we conducted targeted ablation experiments. Table~\ref{tab:dropresults} presents configurations labeled ``Semantic Confidence Drop'' and ``Learnable Structure Drop,'' using only semantic confidence-based or learnable structural drop, respectively. SAD-Splat integrates both mechanisms.

Results show that Learnable Structure Drop achieves an mIoU of 64.01\%, while Semantic Confidence Drop yields an mIoU of 63.79\%. The former uses $L_0$ regularization and Hard Concrete distribution to reduce redundant points, but may discard critical information. The latter relies on a semantic confidence map from SAM and GeoRSCLIP to assess semantic reliability, yet may overly eliminate valuable low-confidence points. In contrast, SAD-Splat, which combines both, achieves an mIoU of 65.22\% and a precision of 82.74\%, demonstrating the complementarity of the submodules: structural drop enhances sparsity, while semantic drop targets ambiguous regions, significantly improving performance and validating the Gaussian Point Drop design. This synergistic mechanism ensures model robustness in complex scenes, enhancing its potential for 3D-AVS-SS tasks.

\subsection{Parameter Analysis}

\textbf{Impact of Base Drop Rate $P_{\text{base}}$ on Model Performance}

\begin{table}[t!]
\centering
\begin{tabular}{lccc}
\toprule
\textbf{Network} & \textbf{mIoU (\%)} & \textbf{Accuracy (\%)} & \textbf{Gaussian Points}\\
\midrule
baseline & 64.44 & 82.01  & 919,553\\
$P_{\text{base}}$=0.5 & 64.66 & 82.71  & 315,218 \\
$P_{\text{base}}$=1.0 & \textbf{65.22} & \textbf{82.74}  & 162,015\\
$P_{\text{base}}$=1.5 & 64.74 & 82.55  & 87,547\\
$P_{\text{base}}$=2.0 & 63.83 & 82.38  & 48,960\\
\bottomrule
\end{tabular}
\caption{Influence of the Base Drop Rate $P_{base}$ on Model Performance. Best results are highlighted in bold.}
\label{tab:gaussianresults}
\end{table}

We conducted a parameter analysis to investigate the impact of the base drop rate $P_{\text{base}}$ on the performance and model sparsity of SAD-Splat. As shown in Table~\ref{tab:gaussianresults}, as $P_{\text{base}}$ increases, the number of Gaussian points gradually decreases, thereby reducing the model size (from 315,218 points at $P_{\text{base}}=0.5$ to 48,760 points at $P_{\text{base}}=2.0$).

When $P_{\text{base}}=1.0$, the model achieves optimal performance, attaining the highest mIoU (65.22\%) and accuracy (82.74\%) with only 162,015 Gaussian points, significantly fewer than the baseline's 919,553 points. At $P_{\text{base}}=0.5$, the model substantially reduces the number of Gaussian points compared to the baseline, while achieving a modest improvement in accuracy (82.71\%) and a slight increase in mIoU (64.66\%). When $P_{\text{base}} \geq 1.5$, segmentation performance noticeably declines, indicating that an excessive drop impairs performance. These results suggest that $P_{\text{base}}=1.0$ strikes the best balance between segmentation accuracy and model compactness.
\section{Conclusion}

In this paper, we propose SAD-Splat, a novel approach for 3D-AVS-SS that addresses the challenges of semantic ambiguity and structural redundancy. We introduce a Gaussian point drop module, which integrates semantic confidence estimation with a learnable sparsity mechanism based on the Hard Concrete distribution. Additionally, we design a high-confidence pseudo-label generation pipeline that effectively leverages 2D foundation models to enhance supervision under limited ground-truth labels. To support this task, we construct a challenging dataset, 3D-AS, encompassing diverse and complex real-world aerial scenes. Experimental results demonstrate that SAD-Splat significantly reduces the number of Gaussian points in the final model while improving segmentation performance, achieving a compact yet expressive 3D representation. Our approach offers a promising direction for efficient and scalable 3D semantic understanding in remote sensing applications.

\newpage

\section*{Reproduction Checklist}
\subsection*{Methodology Description}
\begin{itemize}
    \item Includes a conceptual outline and/or pseudocode description of AI methods introduced. \hfill (\textit{yes})
    \item Clearly delineates statements that are opinions, hypothesis, and speculation from objective facts and results. \hfill (\textit{yes })
    \item Provides well marked pedagogical references for less-familiar readers to gain background necessary to replicate the paper. \hfill (\textit{yes})
\end{itemize}

\subsection*{Theoretical Contributions}
\begin{itemize}
    \item Does this paper make theoretical contributions? \hfill (\textit{yes}) \\
    \textit{If yes, please complete the list below.}
    \begin{itemize}
        \item All assumptions and restrictions are stated clearly and formally. \hfill (\textit{yes})
        \item All novel claims are stated formally (e.g., in theorem statements). \hfill (\textit{yes})
        \item Proofs of all novel claims are included. \hfill (\textit{yes})
        \item Proof sketches or intuitions are given for complex and/or novel results. \hfill (\textit{yes})
        \item Appropriate citations to theoretical tools used are given. \hfill (\textit{yes})
        \item All theoretical claims are demonstrated empirically to hold. \hfill (\textit{NA})
        \item All experimental code used to eliminate or disprove claims is included. \hfill (\textit{NA})
    \end{itemize}
\end{itemize}

\subsection*{Datasets}
\begin{itemize}
    \item Does this paper rely on one or more datasets? \hfill (\textit{yes}) \\
    \textit{If yes, please complete the list below.}
    \begin{itemize}
        \item A motivation is given for why the experiments are conducted on the selected datasets. \hfill (\textit{yes})
        \item All novel datasets introduced in this paper are included in a data appendix. \hfill (\textit{yes})
        \item All novel datasets introduced in this paper will be made publicly available upon publication of the paper with a license that allows free usage for research purposes. \hfill (\textit{yes})
        \item All datasets drawn from the existing literature are accompanied by appropriate citations. \hfill (\textit{NA})
        \item All datasets drawn from the existing literature are publicly available. \hfill (\textit{NA})
        \item All datasets that are not publicly available are described in detail, with explanation why publicly available alternatives are not scientifically satisficing. \hfill (\textit{NA})
    \end{itemize}
\end{itemize}

\subsection*{Computational Experiments}
\begin{itemize}
    \item Does this paper include computational experiments? \hfill (\textit{yes / no}) \\
    \textit{If yes, please complete the list below.}
    \begin{itemize}
        \item This paper states the number and range of values tried per (hyper-)parameter during development, along with the criterion used for selecting the final parameter setting. \hfill (\textit{partial})
        \item Any code required for pre-processing data is included in the appendix. \hfill (\textit{partial})
        \item All source code required for conducting and analyzing the experiments is included in a code appendix. \hfill (\textit{yes})
        \item All source code required for conducting and analyzing the experiments will be made publicly available upon publication of the paper with a license that allows free usage for research purposes. \hfill (\textit{yes})
        \item All source code implementing new methods have comments detailing the implementation, with references to the paper where each step comes from. \hfill (\textit{yes})
        \item If an algorithm depends on randomness, then the method used for setting seeds is described in a way sufficient to allow replication of results. \hfill (\textit{yes})
        \item This paper specifies the computing infrastructure used for running experiments (hardware and software). \hfill (\textit{yes})
        \item This paper formally describes evaluation metrics used and explains the motivation for choosing these metrics. \hfill (\textit{yes})
        \item This paper states the number of algorithm runs used to compute each reported result. \hfill (\textit{yes})
        \item Analysis of experiments goes beyond single-dimensional summaries of performance to include measures of variation, confidence, or other distributional information. \hfill (\textit{no})
        \item The significance of any improvement or decrease in performance is judged using appropriate statistical tests (e.g., Wilcoxon signed-rank). \hfill (\textit{no})
        \item This paper lists all final (hyper-)parameters used for each model/algorithm in the paper’s experiments. \hfill (\textit{yes})
    \end{itemize}
\end{itemize}


\begin{thebibliography}{20}
\providecommand{\natexlab}[1]{#1}

\bibitem[{Caron et~al.(2021)Caron, Touvron, Misra, J{\'e}gou, Mairal, Bojanowski, and Joulin}]{caron2021emerging}
Caron, M.; Touvron, H.; Misra, I.; J{\'e}gou, H.; Mairal, J.; Bojanowski, P.; and Joulin, A. 2021.
\newblock Emerging properties in self-supervised vision transformers.
\newblock In \emph{Proceedings of the IEEE/CVF international conference on computer vision}, 9650--9660.

\bibitem[{Cen et~al.(2025)Cen, Fang, Yang, Xie, Zhang, Shen, and Tian}]{cen2025segment}
Cen, J.; Fang, J.; Yang, C.; Xie, L.; Zhang, X.; Shen, W.; and Tian, Q. 2025.
\newblock Segment any 3d gaussians.
\newblock In \emph{Proceedings of the AAAI Conference on Artificial Intelligence}, volume~39, 1971--1979.

\bibitem[{Cherti et~al.(2023)Cherti, Beaumont, Wightman, Wortsman, Ilharco, Gordon, Schuhmann, Schmidt, and Jitsev}]{cherti2023reproducible}
Cherti, M.; Beaumont, R.; Wightman, R.; Wortsman, M.; Ilharco, G.; Gordon, C.; Schuhmann, C.; Schmidt, L.; and Jitsev, J. 2023.
\newblock Reproducible scaling laws for contrastive language-image learning.
\newblock In \emph{Proceedings of the IEEE/CVF conference on computer vision and pattern recognition}, 2818--2829.

\bibitem[{Dong et~al.(2023)Dong, Bao, Zheng, Zhang, Chen, Yang, Zeng, Zhang, Yuan, Chen et~al.}]{dong2023maskclip}
Dong, X.; Bao, J.; Zheng, Y.; Zhang, T.; Chen, D.; Yang, H.; Zeng, M.; Zhang, W.; Yuan, L.; Chen, D.; et~al. 2023.
\newblock Maskclip: Masked self-distillation advances contrastive language-image pretraining.
\newblock In \emph{Proceedings of the IEEE/CVF Conference on Computer Vision and Pattern Recognition}, 10995--11005.

\bibitem[{Fu et~al.(2024)Fu, Hamilton, Brandt, Feldman, Zhang, and Freeman}]{fu2024featup}
Fu, S.; Hamilton, M.; Brandt, L.; Feldman, A.; Zhang, Z.; and Freeman, W.~T. 2024.
\newblock Featup: A model-agnostic framework for features at any resolution.
\newblock \emph{arXiv preprint arXiv:2403.10516}.

\bibitem[{Huang et~al.(2023)Huang, Jiang, Lv, Liu, and Fu}]{huang2023deep}
Huang, L.; Jiang, B.; Lv, S.; Liu, Y.; and Fu, Y. 2023.
\newblock Deep-learning-based semantic segmentation of remote sensing images: A survey.
\newblock \emph{IEEE Journal of Selected Topics in Applied Earth Observations and Remote Sensing}, 17: 8370--8396.

\bibitem[{Kerbl et~al.(2023)Kerbl, Kopanas, Leimk{\"u}hler, and Drettakis}]{kerbl20233d}
Kerbl, B.; Kopanas, G.; Leimk{\"u}hler, T.; and Drettakis, G. 2023.
\newblock 3d gaussian splatting for real-time radiance field rendering.
\newblock \emph{ACM Trans. Graph.}, 42(4): 139--1.

\bibitem[{Kerr et~al.(2023)Kerr, Kim, Goldberg, Kanazawa, and Tancik}]{kerr2023lerf}
Kerr, J.; Kim, C.~M.; Goldberg, K.; Kanazawa, A.; and Tancik, M. 2023.
\newblock Lerf: Language embedded radiance fields.
\newblock In \emph{Proceedings of the IEEE/CVF international conference on computer vision}, 19729--19739.

\bibitem[{Kirillov et~al.(2023)Kirillov, Mintun, Ravi, Mao, Rolland, Gustafson, Xiao, Whitehead, Berg, Lo et~al.}]{kirillov2023segment}
Kirillov, A.; Mintun, E.; Ravi, N.; Mao, H.; Rolland, C.; Gustafson, L.; Xiao, T.; Whitehead, S.; Berg, A.~C.; Lo, W.-Y.; et~al. 2023.
\newblock Segment anything.
\newblock In \emph{Proceedings of the IEEE/CVF international conference on computer vision}, 4015--4026.

\bibitem[{Li et~al.(2022)Li, Weinberger, Belongie, Koltun, and Ranftl}]{li2022language}
Li, B.; Weinberger, K.~Q.; Belongie, S.; Koltun, V.; and Ranftl, R. 2022.
\newblock Language-driven semantic segmentation.
\newblock \emph{arXiv preprint arXiv:2201.03546}.

\bibitem[{Liu et~al.(2023)Liu, Zhan, Zhang, Xu, Yu, El~Saddik, Theobalt, Xing, and Lu}]{liu2023weakly}
Liu, K.; Zhan, F.; Zhang, J.; Xu, M.; Yu, Y.; El~Saddik, A.; Theobalt, C.; Xing, E.; and Lu, S. 2023.
\newblock Weakly supervised 3d open-vocabulary segmentation.
\newblock \emph{Advances in Neural Information Processing Systems}, 36: 53433--53456.

\bibitem[{Louizos, Welling, and Kingma(2017)}]{louizos2017learning}
Louizos, C.; Welling, M.; and Kingma, D.~P. 2017.
\newblock Learning sparse neural networks through $ L\_0 $ regularization.
\newblock \emph{arXiv preprint arXiv:1712.01312}.

\bibitem[{Mildenhall et~al.(2021)Mildenhall, Srinivasan, Tancik, Barron, Ramamoorthi, and Ng}]{mildenhall2021nerf}
Mildenhall, B.; Srinivasan, P.~P.; Tancik, M.; Barron, J.~T.; Ramamoorthi, R.; and Ng, R. 2021.
\newblock Nerf: Representing scenes as neural radiance fields for view synthesis.
\newblock \emph{Communications of the ACM}, 65(1): 99--106.

\bibitem[{Piekenbrinck et~al.(2025)Piekenbrinck, Schmidt, Hermans, Vaskevicius, Linder, and Leibe}]{piekenbrinck2025opensplat3d}
Piekenbrinck, J.; Schmidt, C.; Hermans, A.; Vaskevicius, N.; Linder, T.; and Leibe, B. 2025.
\newblock OpenSplat3D: Open-Vocabulary 3D Instance Segmentation using Gaussian Splatting.
\newblock In \emph{Proceedings of the Computer Vision and Pattern Recognition Conference}, 5246--5255.

\bibitem[{Qin et~al.(2024)Qin, Li, Zhou, Wang, and Pfister}]{qin2024langsplat}
Qin, M.; Li, W.; Zhou, J.; Wang, H.; and Pfister, H. 2024.
\newblock Langsplat: 3d language gaussian splatting.
\newblock In \emph{Proceedings of the IEEE/CVF Conference on Computer Vision and Pattern Recognition}, 20051--20060.

\bibitem[{Radford et~al.(2021)Radford, Kim, Hallacy, Ramesh, Goh, Agarwal, Sastry, Askell, Mishkin, Clark et~al.}]{radford2021learning}
Radford, A.; Kim, J.~W.; Hallacy, C.; Ramesh, A.; Goh, G.; Agarwal, S.; Sastry, G.; Askell, A.; Mishkin, P.; Clark, J.; et~al. 2021.
\newblock Learning transferable visual models from natural language supervision.
\newblock In \emph{International conference on machine learning}, 8748--8763. PmLR.

\bibitem[{Rahnemoonfar, Chowdhury, and Murphy(2022)}]{rahnemoonfar2022rescuenet}
Rahnemoonfar, M.; Chowdhury, T.; and Murphy, R. 2022.
\newblock RescueNet: A high resolution UAV semantic segmentation benchmark dataset for natural disaster damage assessment.
\newblock \emph{arXiv preprint arXiv:2202.12361}.

\bibitem[{Ravi et~al.(2024)Ravi, Gabeur, Hu, Hu, Ryali, Ma, Khedr, R{\"a}dle, Rolland, Gustafson et~al.}]{ravi2024sam}
Ravi, N.; Gabeur, V.; Hu, Y.-T.; Hu, R.; Ryali, C.; Ma, T.; Khedr, H.; R{\"a}dle, R.; Rolland, C.; Gustafson, L.; et~al. 2024.
\newblock Sam 2: Segment anything in images and videos.
\newblock \emph{arXiv preprint arXiv:2408.00714}.

\bibitem[{Zhang et~al.(2024)Zhang, Zhao, Guo, and Yin}]{zhang2024rs5m}
Zhang, Z.; Zhao, T.; Guo, Y.; and Yin, J. 2024.
\newblock RS5M and GeoRSCLIP: A large scale vision-language dataset and a large vision-language model for remote sensing.
\newblock \emph{IEEE Transactions on Geoscience and Remote Sensing}.

\bibitem[{Zhou et~al.(2024)Zhou, Chang, Jiang, Fan, Zhu, Xu, Chari, You, Wang, and Kadambi}]{zhou2024feature}
Zhou, S.; Chang, H.; Jiang, S.; Fan, Z.; Zhu, Z.; Xu, D.; Chari, P.; You, S.; Wang, Z.; and Kadambi, A. 2024.
\newblock Feature 3dgs: Supercharging 3d gaussian splatting to enable distilled feature fields.
\newblock In \emph{Proceedings of the IEEE/CVF Conference on Computer Vision and Pattern Recognition}, 21676--21685.

\end{thebibliography}
\end{document}